# Terminologies augmented recurrent neural network model for clinical named entity recognition

Ivan Lerner, MSc[1-2]; Nicolas Paris, MSc[2-3]; Xavier Tannier, PhD[4]

[1] Paris University, Paris, France

[2] AP-HP, DSI-WIND, Paris, France

[3] LIMSI, CNRS, Univ. Paris-Sud, Université Paris-Saclay, F-91405, Orsay, France

[4] Sorbonne Université, Inserm, Univ Paris 13, Laboratoire d'Informatique Médicale et d'Ingénierie des Connaissances pour la e-Santé, LIMICS, F-93017 Bobigny, France

**Corresponding author:** Xavier Tannier

Sorbonne Université, Inserm, Univ Paris 13, Laboratoire d'Informatique Médicale et d'Ingénierie des Connaissances pour la e-Santé, LIMICS, F-93017 Bobigny, France
E-mail: xavier.tannier@sorbonne-universite.fr

# Abstract

## Objective

We aimed to enhance the performance of a supervised model for clinical named-entity recognition (NER) using medical terminologies. In order to evaluate our system in French, we built a corpus for 5 types of clinical entities.

## Methods

We used a terminology-based system as baseline, built upon UMLS and SNOMED. Then, we evaluated a biGRU-CRF, and an hybrid system using the prediction of the terminology-based system as feature for the biGRU-CRF. In English, we evaluated the NER systems on the i2b2-2009 Medication Challenge for Drug name recognition, which contained 8,573 entities for 268 documents. In French, we built APcNER, a corpus of 147 documents annotated for 5 entities (drug name, sign or symptom, disease or disorder, diagnostic procedure or lab test and therapeutic procedure). We evaluated each NER systems using exact and partial match definition of F-measure for NER.

## Results

The APcNER contains 4,837 entities which took 28 hours to annotate, the inter-annotator agreement was acceptable for Drug name in exact match (85%) and acceptable for other entity types in non-exact match (>70%). For drug name recognition on both i2b2-2009 and APcNER, the biGRU-CRF performed better that the terminology-based system, with an exact-match F-measure of 91.1% versus 73% and 81.9% versus 75% respectively. Moreover, the hybrid system outperformed the biGRU-CRF, with an exact-match F-measure of 92.2% versus 91.1% (i2b2-2009) and 88.4% versus 81.9% (APcNER). On APcNER corpus, the micro-average F-measure of the hybrid system on the 5 entities was 69.5% in exact match, and 84.1% in non-exact match.

## Conclusion

APcNER is a French corpus for clinical-NER of five type of entities which covers a large variety of document types. Extending supervised model with terminology allowed for an easy performance gain, especially in low regimes of entities, and established near state of the art results on the i2b2-2009 corpus.

Highlights

- For 28 hours of annotation time, we built APcNER, a French corpus for clinical named-entity recognition which covers a large variety of document types
- APcNER allowed to achieve an average 84% non-exact F-measure on five type of clinical entities
- We provide consistent results on English and French corpora that give insight into the complementarity of a terminology with a supervised model

# Introduction

Within the range of data covered by electronic health records (EHRs), clinical documents (e. g. discharge summaries or physicians' letters) are rich sources of information for various applications such as patient recruitment for clinical research, epidemiological surveillance, medical coding and decision-making tools [1]. Information extraction tools must be tailored for application in the medical domain, where the language is both unstructured (e. g. free text) and semi-structured (e. g. drug lists), with a wide vocabulary.

Named-entity recognition (NER) is the mention detection and type classification of named entities, where named entities are concepts that can be referenced by various linguistic expressions. In recent decades, there has been a growing interest in clinical-NER, the task of NER for medical concepts such as drug name, disease or signs [2]. Supervised systems based on machine learning have proven to be more efficient than rule-based and terminology-based systems for NER[3]. Research efforts have then been made to unify these methods in hybrid systems, in a purely unsupervised [4,5] or semi-supervised fashion [6–8]. Such approaches are motivated by the necessity to reduce the need for manually annotated examples in the case of a supervised system, or the need for handwritten rules by experts in the field of rule-based and ontological systems.

In addition, for languages other than English, annotated corpora and ontologies are scarcer. For instance, in French, there is only one annotated clinical corpus which covers a small subset of the medical domain [9], and international ontologies such as the Unified Medical Language System (UMLS®) are not fully translated [10]. The development cost of such annotated corpus is very high, as it has been reported that annotating 5 medical documents for 12 entities take on average 82 mins [9].

In this study, we aimed to evaluate a clinical-NER system which development could scale up to the different uses of a large French data warehouse [11]. First, we built an annotated corpus for five clinical entity types and present the details of the annotation process. Second, we evaluated three different systems: 1/ a terminology-based system built upon the Unstructured Information Management Architecture (apache-UIMA®) framework, 2/ a supervised neural model based on a biGRU-CRF architecture, and 3/ an hybrid system. These experiments were achieved on our French corpus as well as on a well-known, freely available corpus in English, for comparison purposes.

# Methods

## Corpus and annotation process

We used two corpora in this study, an English corpus from the i2b2-2009 Medication challenge (i2b2-2009) [12] and a French corpus APcNER, with clinical reports extracted from the AP-HP data warehouse [11].

### i2b2-2009 corpus

The original corpus included 1243 de-identified discharge summaries, 17 of which were annotated by the i2b2 team, and 251 collectively annotated by the challenge participants.[1] Overall, the 268 annotated documents of the corpus contained 337,745 tokens, 8,573 entities and 17,933 sentences, for a vocabulary of 23,214 tokens. The median sentence length was 13. The overall 8,573 entities, comprised 6,488 (75%) unigrams, 1,053 (12%) bigrams and 1,029 (12%) longer mentions. These 268 discharge summaries were randomly assigned to a train (70%), development (15%) and test set (15%). We only kept the name of the drug from the annotations and discarded the dose and other drug-related information.

We also used a randomly sampled subset of the i2b2-2009 train set, to include the same number of drug name entities than the APcNER corpus, for quantitative comparison purposes. We call this smaller corpus "i2b2-small".

### APcNER corpus

We randomly sampled (stratified on document types) 147 documents from the dataset used for de-identification at AP-HP, excluding prescriptions and admission reports. The AP-HP de-identification dataset is a set of 3223 French-language medical documents sampled - with upsampling of rare documents types - over 50 millions documents from the AP-HP Data Warehouse, which included EHR data from 39 hospitals. The APcNER included 4 main types of documents: discharge summaries, letters from physicians, operating reports and additional examination reports. Detailed document types can be found in the supplementary Table S1. We based our annotation guideline on the UMLS® semantic types [13], Table 1 details the 5 medical entities that we annotated (drug name, sign or symptom, disease or disorder, diagnostic procedure or lab test and therapeutic procedure). We used BRAT Rapid Annotation Tool (BRAT) [14]. The general guidance for annotation was to annotate the most complete entities (e. g. "Non-ST segment elevation myocardial infarction" and not only "myocardial infarction") with no possible overlap between entities (see Appendix Section 2 for the detailed annotation guideline). Documents were annotated by groups of 10 and the annotation time monitored. Documents were pre-annotated using a terminology-based annotator (see below). IL, a medical resident, annotated all the dataset. Then, in order to estimate the quality of the guidelines, we randomly selected 10 documents that NG, a medical doctor, annotated blindly. We then assessed the agreement between the two annotators. Conflicting annotations have been discussed between IL and NG and are referenced in the Annotation Guideline (see Appendix Section 2). IL went through all the documents a second time in order to disambiguate the conflicting annotations. Finally, for homogenization purposes, we fitted a simple conditional random field (CRF) model to the dataset with the default NER features using Wapiti [15]. We used this model to detect annotation inconsistencies or case errors, and manually corrected them. Finally, we randomly divided all the documents into 6 folds by stratifying on the types of documents and the length of the documents. The corpus is made available on condition that a research project is submitted to the scientific and ethics committee of the AP-HP health data warehouse (https://recherche.aphp.fr/eds/recherche/).

---

[1] Note that the original test set used during the i2b2 challenge is made of the 251 collectively annotated documents.

## Clinical-NER systems

For all experiments, we used the inside, outside, beginning (IOB) tagging scheme [16]. For an entity of type DRUG, the first token of such entity is coded B-DRUG, if the entity is constituted of multiples tokens, the following tokens are coded I-DRUG, and all tokens outside entities are coded O. For instance, "placed on heparin sodium" is encoded "O O B-DRUG I-DRUG".

| Entity types | Semantic Type | UMLS Semantic Tree Number | Number of terms (All sources) |
|---|---|---|---|
| Drug name | Antibiotic<br>Clinical Drug<br>Pharmacologic Substance<br>Vitamin | A1.4.1.1.1.1<br>A1.3.3<br>A1.4.1.1.1<br>A1.4.1.1.3.4 | French: 24,932<br>English: 96,547 |
| Sign or symptom | Sign or Symptom | A2.2.2 | 5,125 |
| Disease or disorder | Mental or Behavioral Dysfunction<br>Cell or Molecular Dysfunction<br>Anatomical Abnormality<br>Congenital Abnormality<br>Acquired Abnormality<br>Injury or Poisoning<br>Pathologic Function<br>Neoplastic Process<br>Disease or Syndrome | B2.2.1.2.1.1<br>B2.2.1.2.2<br>A1.2.2<br>A1.2.2.1<br>A1.2.2.2<br>B2.3<br>B2.2.1.2<br>B2.2.1.2.1.2<br>B2.2.1.2.1 | 104,104 |
| Diagnostic procedure or lab test | Laboratory or Test Result<br>Laboratory Procedure<br>Diagnostic Procedure | A2.2.1<br>B1.3.1.1<br>B1.3.1.2 | 16,974 |
| Therapeutic procedure | Therapeutic or Preventive Procedure | B1.3.1.3 | 20,926 |

**Table 1. UMLS semantic types extracted for each entity**

## Terminology based system

In English, we extracted drug names using regular expressions from UMLS® (including SNOMED 3.5 CT®) to create a large dictionary of drug names. In French, we used 10 terminologies, 8 of which were previously referenced in [10] (ATC, BPDM, CCAM, CIM-10, DRC, SNOMED, UMLS), and 2 terminologies held by AP-HP (GLIMS, QDOC). Table 1 details the extracted UMLS semantic types by entity type. Terms were extracted using minimal regular expression rules, then tokenized using Stanford CoreNLP [17]. We discarded common terms based on Wikipedia word count. The matching rules were based on the apache-UIMA framework, CoreNLP and dkPRO and allowed multiple word matching, stop words, accent normalization and case insensitive matching. The source code is available with the GLP-3 license (https://github.com/EDS-APHP/uima-aphp/tree/master/uima-dict). Resolution of conflicting (overlapping) entities was done by randomly picking one of the conflicting entities.

### Supervised system

Sentence segmentation and tokenization were performed using Stanford CoreNLP [17]. Numbers were normalized to a unique token. We learned a biGRU-CRF (Bidirectional - Gated Recurrent Unit - Conditional Random Field) [3], based on the NCRF++ implementation [18]. The model takes 2 types of inputs. First, word embeddings trained with the Skip-Gram model [19], 2 millions AP-HP documents for French (dimension 200), and 2 millions MIMIC [20] clinical notes for English (dimension 100). Second, character embeddings, which are processed by 1-dimensional convolution (kernel size 3) with max-pooling. The global token representation is the concatenation of the word and character embeddings (see Figure 1). The sequence of token representation is then processed forward and backward by the biGRU, which outputs a emission probability score for each entity class. Finally, the CRF decodes the sequence of labels by associating the emission probability score with a transition probability score (see Figure 2). We used a dropout rate of 0.5, an L2-norm on the model weights and early stopping to prevent overfitting. We used Bayesian optimization [21] to perform hyperparameter tuning of the architecture (number of layers, number of neurons, character embedding dimension), learning rate and L2-norm. Note that with 1 entity type, NER is a 3 class classification problem, and with $k$ entity types it is a $k \times 2 + 1$ classification problem, e. g. with the 5 entities type of APcNER we have 11 class to predict.

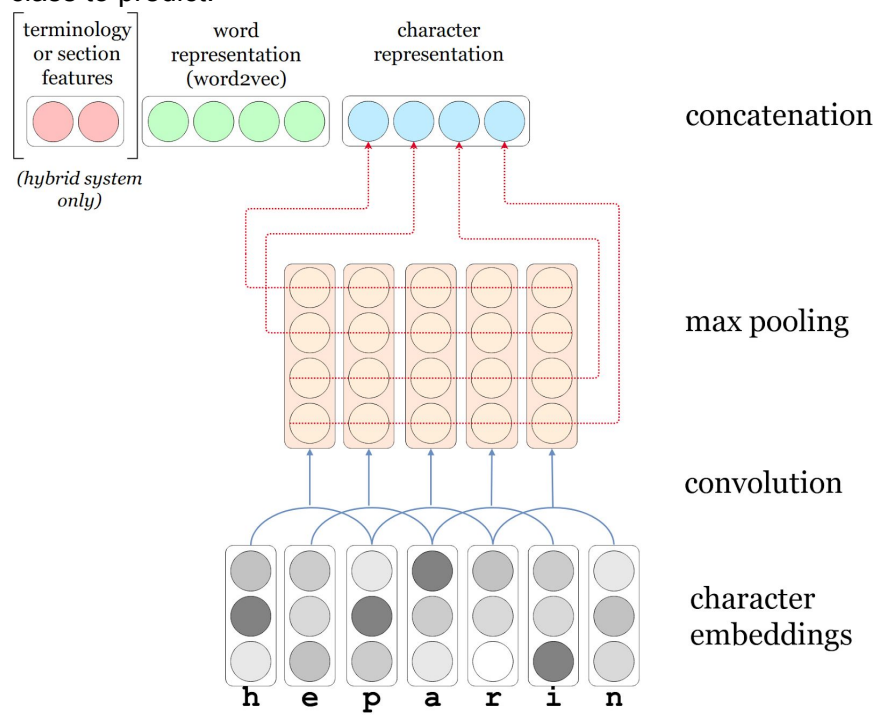

**Figure 1.** Word representation

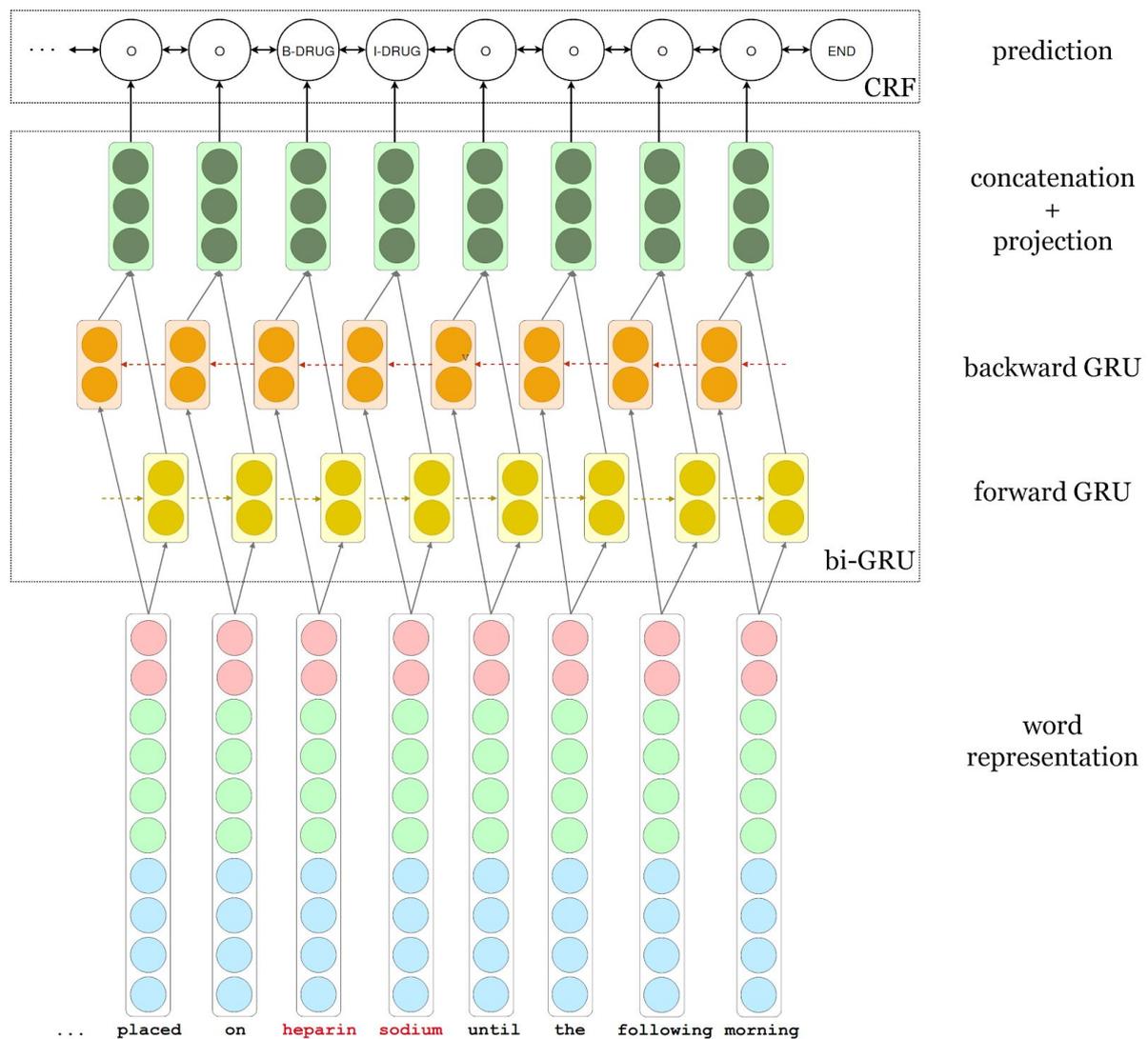
**Figure 2.** biGRU-CRF architecture

Hybrid system

We proposed a hybrid system in which a supervised model was associated with a terminology-based model. For each token, we added a feature representing the class predicted by the terminology-based system described above, which is then encoded as a categorical embedding of dimension 5. This embedding is then concatenated to the word embedding from the supervised system. This terminology based feature can take two values per entity types (e. g. B-Drug Name; I-Drug name), as well as one value for the "Outside" class. We also added a context feature based on terminologies of sections headings that we developed internally. The French section headings terminology was created based on documents of the same distribution as the APcNER corpus, the English section heading terminology was created for the 2018 AP-HP Datathon based on MIMIC notes. For each token, the context feature was the class of the last section heading, following the order of the document. Then, the context feature is encoded as a categorical embedding of dimension 5. This embedding is concatenated to the word embedding from the supervised system and the other feature embedding.

## Evaluation methodology and metrics

First, we evaluated the systems on the i2b2-2009 corpus and its reduced version, i2b2-small, and on the APcNER corpus with labels limited to drug names. Then we evaluated the systems on the entire APcNER corpus, as one multi-class task. We compared the models based on F-measure, precision and recall using the CONLL definition: "*precision is the percentage of named entities found by the learning system that are correct. Recall is the percentage of named entities present in the corpus that are found by the system. A named entity is correct only if it is an exact match of the corresponding entity in the data file*". We also compared the model based on partial match, which allowed the boundaries of the entity to mismatch.

Neural network models training is highly non-deterministic and is subject to the random seed choice. Because of this variability during the training phase, we performed five experiments for each model presented in this work, and reported the mean, minimum and maximum of each metric.

For the i2b2-2009 corpus, we selected the optimal set of hyperparameters for the supervised model based on a development set, including the optimal epoch stop, and evaluated on the test set the models trained on the train+dev set. For *APcNER*, we selected the optimal set of hyperparameters for the supervised model by 6 fold cross-validation, we defined the optimal epoch stop as the mean of the optimal epoch stop for each fold. We evaluate the model on each fold, after training it on the remaining 5 folds, using the same set of optimal hyperparameters. We then report the evaluation metrics computed over the 6 folds.

# Results

## Corpus and annotation process

### APcNER corpus

The first round took in average 87 min per 10 documents and a total of 22 hours. The second round took in average 28 min per 10 documents and a total of 7 hours. Both round took in average 115 minutes per 10 documents and a total of 28 hours. The inter-annotator agreement after the first two round and before the CRF harmonization is reported in Table 2. For drug names, the inter-annotator agreement is acceptable (F-measure .85) and good considering non-exact match (F-measure .92). Overall, the 147 documents of the corpus contain 80,421 tokens, 4,837 entities and 3,093 sentences, for a vocabulary of 12,523 tokens. The median sentence length is 14.

| Entity types | Non exact F-measure | Exact F-measure | # entities | n-grams (%) | | |
|---|---|---|---|---|---|---|
| | | | | n = 1 | n = 2 | n ≥ 3 |
| **Drug name** | .92 | .85 | 1076 | 1014 (94) | 54 (.5) | 8 (.1) |
| **Sign or symptom** | .71 | .55 | 432 | 356 (82) | 65 (15) | 11 (.3) |
| **Disease or disorder** | .77 | .65 | 1672 | 1238 (74) | 330 (20) | 104 (.6) |
| **Diagnostic procedure or lab test** | .87 | .70 | 1156 | 808 (70) | 297 (27) | 51 (.4) |
| **Therapeutic procedure** | .71 | .51 | 501 | 414 (83) | 73 (15) | 14 (.3) |

**Table 2. APcNER inter-annotator agreement.**
The F-measure is the harmonic mean between precision and recall computed as in Conll 2003. Inter-annotator agreement is evaluated on a random subset of APcNER of 10 documents.

## Clinical-NER systems

| Corpus | System | Exact-match | | | Partial-match | | |
| --- | --- | --- | --- | --- | --- | --- | --- |
| | | F[a] [min-max] | P[b] [min-max] | R[c] [min-max] | F[a] [min-max] | P[b] [min-max] | R[c] [min-max] |
| i2b2-2009 | Terminologies | 73.0 | 76.7 | 69.7 | 84.6 | 88.9 | 80.6 |
| | biGRU-CRF | 91.1 [90.0-91.8] | 90.6 [89.7-92.6] | 91.7 [87.6-93.1] | 93.5 [92.4-94.1] | 92.9 [92.2-95.0] | 94.2 [90.0-96.0] |
| | Hybrid system | **92.2** [91.2-93.0] | 92.1 [91.4-93.1] | 92.2 [90.5-93.8] | **94.7** [94.3-95.2] | 94.6 [94.0-95.2] | 94.7 [93.5-96.1] |
| i2b2-small | biGRU-CRF | 85.6 [84.7-86.2] | 85.2 [83.0-87.2] | 86.0 [84.2-89.6] | 90.4 [89.7-91.2] | 90.1 [87.2-92.0] | 90.8 [88.6-94.1] |
| | Hybrid system | **87.8** [85.9-88.8] | 88.2 [86.8-89.4] | 87.4 [83.7-89.7] | **90.6** [85.1-93.0] | 92.0 [90.8-93.4] | 89.4 [80.1-93.8] |
| APcNER | Terminologies | 75.0 | 70.8 | 79.7 | 77.7 | 73.3 | 82.5 |
| | biGRU-CRF | 81.9 [81.2-82.4] | 86.6 [84.9-88.7] | 77.8 [76.6-78.9] | 86.4 [85.1-87.7] | 91.4 [90.3-93.4] | 82.0 [80.1-84.0] |
| | Hybrid system | **86.4** [86.2-86.8] | 89.6 [87.7-90.9] | 83.4 [82.3-84.7] | **90.4** [89.9-91.1] | 93.8 [91.9-94.9] | 87.2 [85.9-88.6] |

[a]F-measure; [b]Precision; [c]Recall

**Table 3. Drug name recognition.**
Comparison between a terminology-based system, a supervised model (biGRU-CRF) and a hybrid system on the i2b2-2009 corpus, i2b2-small (a reduced version of the former corpus), and APcNER. The evaluation metrics are F-measure, precision and recall for exact match and partial match.

Table 3 summarizes the results of Drug name recognition in the i2b2-2009 corpus, i2b2-small (reduced version of the former dataset) and the APcNER corpus. For both i2b2-2009 and APcNER, the biGRU-CRF outperforms the terminology-based system, with an exact-match F-measure of 91.1% versus 73% and 81.9% versus 75% respectively. For both i2b2-2009 and APcNER, the hybrid system outperforms the biGRU-CRF, with an exact-match F-measure of 92.2% versus 91.1% and 88.4% versus 81.9% respectively. The performance on i2b2-small is very close to the performance on APcNER for the hybrid system (with an exact-match F-measure of 87.8% and 86.4%).

Table 4 summarises the results of clinical-NER on all the entity types of the APcNER corpus. For all three systems, the exact-match performance for Sign or symptom, Disease or disorder, Diagnostic procedure or lab test, Therapeutic procedure are much lower compared to Drug name. The difference between the biGRU-CRF and the terminologies are also more important than for Drug names, with exact-match F-measures of 55.2% versus 15.7% (Sign or symptom), 59.5% versus 30.9% (Disease or disorder), 75.9% versus 30.4% (Diagnostic procedure or lab test) and 61.3% versus 16.6% (Therapeutic procedure). The difference between exact-match and partial match metrics is also greater than for Drug names. The hybrid system outperforms the other systems for all entity types except Therapeutic procedure. Table S3 summarizes the hyperparameters of the models.

| Corpus | System | Exact-match | | | Partial-match | | |
| --- | --- | --- | --- | --- | --- | --- | --- |
| | | F[a] [min-max] | P[b] [min-max] | R[c] [min-max] | F[a] [min-max] | P[b] [min-max] | R[c] [min-max] |
| All types (micro-average) | Terminologies | 32.5 | 49.2 | 24.3 | 48.4 | 73.5 | 36.1 |
| | biGRU-CRF | 67.1 [66.8-67.6] | 69.1 [68.3-69.4] | 65.2 [64.4-66.1] | 82.9 [82.8-83.2] | 85.0 [84.4-86.0] | 80.9 [80.7-81.3] |
| | Hybrid system | **69.5** [69.1-69.7] | 71.6 [71.1-72.2] | 67.5 [67.2-68.2] | **84.1** [84.0-84.3] | 86.3 [85.8-86.8] | 82.1 [81.4-82.6] |
| Drug name | Terminologies | 75.0 | 70.8 | 79.7 | 77.7 | 73.3 | 82.5 |
| | biGRU-CRF | 81.5 [80.2-82.7] | 84.6 [81.9-87.9] | 78.6 [78.1-79.8] | 85.9 [85.1-86.8] | 89.2 [87.2-92.3] | 82.8 [82.0-83.9] |
| | Hybrid system | **85.9** [85.2-86.7] | 87.3 [86.7-88.6] | 84.5 [83.4-85.4] | **90.4** [89.9-91.0] | 92.0 [90.6-92.7] | 89.0 [87.9-89.6] |
| Sign or symptom | Terminologies | 15.7 | 38.9 | 9.8 | 28.4 | 70.4 | 17.8 |
| | biGRU-CRF | 55.2 [53.5-56.3] | 56.6 [55.4-57.6] | 53.8 [51.1-56.2] | 76.5 [75.4-77.2] | 77.4 [76.6-78.6] | 75.6 [74.1-77.8] |
| | Hybrid system | **59.9** [59.6-60.1] | 61.6 [59.8-63.3] | 58.3 [57.2-59.3] | **78.8** [78.2-79.4] | 80.3 [78.7-82.3] | 77.4 [75.8-79.4] |
| Disease or disorder | Terminologies | 30.9 | 33.3 | 28.9 | 62.5 | 68.1 | 57.8 |
| | biGRU-CRF | 59.5 [58.4-60.9] | 61.9 [60.3-63.5] | 57.4 [56.6-58.6] | 79.4 [78.8-79.8] | 82.2 [81.3-82.9] | 76.9 [76.4-77.9] |
| | Hybrid system | **62.4** [61.2-63.0] | 64.5 [63.1-65.1] | 60.4 [59.3-61.0] | **80.9** [80.6-81.1] | 83.1 [82.9-83.4] | 78.8 [78.0-79.2] |
| Diagnostic procedure or lab test | Terminologies | 30.4 | 61.4 | 20.2 | 40.1 | 81.3 | 26.6 |
| | biGRU-CRF | 75.9 [75.3-76.5] | 77.1 [76.3-78.1] | 74.8 [74.2-75.7] | 88.0 [87.4-88.4] | 89.2 [88.3-89.8] | 86.9 [86.3-87.3] |
| | Hybrid system | **77.7** [77.5-77.8] | 79.0 [78.5-79.4] | 76.4 [75.8-76.6] | **88.5** [88.4-88.7] | 89.8 [89.5-90.1] | 87.3 [87.1-87.9] |
| Therapeutic procedure | Terminologies | 16.6 | 36.1 | 10.8 | 35.1 | 76.6 | 22.8 |
| | biGRU-CRF | **61.3** [60.9-61.8] | 64.4 [63.2-65.9] | 58.5 [57.5-59.4] | **80.9** [80.6-81.4] | 84.6 [83.2-86.3] | 77.5 [76.0-78.4] |
| | Hybrid system | 60.2 [59.7-60.9] | 64.4 [63.4-66.3] | 56.5 [55.2-58.0] | 80.1 [79.4-80.6] | 85.1 [83.6-87.6] | 75.7 [73.6-77.2] |

[a]F-measure; [b]Precision; [c]Recall

**Table 4. Multiclass clinical named entity recognition.**
Comparison between a terminology-based system, a supervised model (biGRU-CRF) and a hybrid system on the APcNER dataset. The evaluation metrics are F-measure, precision and recall for exact match and partial match.

## Discussion

In this study, we built APcNER a corpus for clinical-NER of 5 types of entities (Drug names, Sign or symptom, Disease or disorder, Diagnostic procedure or lab test, Therapeutic procedure). We then systematically evaluated a supervised model (biGRU-CRF) against a terminology-based system. Finally, we proposed to extend the supervised system by encoding the prediction of the terminology-based system as categorical embeddings. On the APcNER and on the i2b2-2009 corpora, the biGRU-CRF outperforms the terminology-based system, and the hybrid system is more efficient except for the Therapeutic procedure class.

Both the biGRU-CRF and its extended version outperforms previous results from the i2b2 2009 Medication Challenge (90% F-measure for the best team) [21]. These results (mean 92.2, range [91.2-93.0]) are very close to FABLE [22] which used *bootstrapping*, a semi-supervised approach, leading to 93% F-measure. As the number of examples increases, the information brought by the terminology should become redundant with the one brought by the annotations, which could explain the relatively larger performance gain of the hybrid system in low regime of trained examples (see Table 3). Note that, as mentioned above, our test set is a sub-sample of the i2b2 test set used during the challenge.

The difference in performance for the biGRU-CRF between the i2b2-2009 and the APcNER corpora is partially explained by their difference in number of annotated entities. Indeed, our results (see Table 3) on the reduced version of i2b2 show close performance with the hybrid system when the training set is reduced to the same number of entities than APcNER. Differences remain with the system using no external resources, but this can come from the fact that the domain covered by the APcNER corpus is much broader in terms of document types and medical specialities. Another noteworthy difference is that a drug name followed by its commercial name between brackets is annotated as a single entity in i2b2-2009, but as several separate entities in APcNER, which explains the difference in the distribution of long n-grams (n ≥ 3) between the two corpus (see Table 2).

Another notable result is that the performance of the biGRU-CRF is much lower for other types of entities than Drug names, and the difference between exact match and partial match is larger (see Table 4). Along with the results of the APcNER annotation process (Table 2), it suggests that it is in part due to the longer size of entities. Indeed, compared to Drug names, the other types of entities have between 3 and 6 times more entities composed of at least 2 tokens. The inter-annotator agreement is also lower for these types of entities, and the results of the biGRU-CRF are consistent with those of the the inter-annotator agreement. This is consistent with feedback from the annotation process that boundaries are more difficult to define for longer entities. In addition, the conflict between overlapping entities rarely concerns Drug names, whereas they are more likely to occur between Diagnostic and Therapeutic procedure (e. g. angiography), or between Disorder and Sign (e. g. hemiparesis). Following this analysis, we argue that for the APcNER corpus, the metric of reference should be the non-exact F-measure for entities other than Drug names.

In comparison with MERLOT [9], which include 44,740 entities of 12 types, for 500 documents from Hepato-gastro-enterology and Nutrition ward, APcNER is both smaller (147 documents) and covering a larger domain (no restriction of medical specialty). The inter-annotator agreements for class common to both corpora are comparable, with Sign and

Symptom exact F-measure 59% versus 55%, Drug names 90% versus 85%, and disorder 77% versus 65% for MERLOT and APcNER, respectively.

To our knowledge, our study is the first to provide, for a distribution of medical documents that is representative of that of an EHR (with the exception of imagery reports, prescriptions and initial observation), an estimation of the annotation cost for clinical-NER, which is achieving on average 84% non-exact match F-measure for 28 hours of annotations. Using active learning is likely to diminish this cost by 40 to 80% [23,24], hence achieving performance greater than 95% on this task seems reachable. In addition, we provide consistent results in English and French, which provide an insight into the complementarity of a terminology with a supervised model.

The main limitation of our study is the small size of our corpus compared to the large domain it covers. However, using cross-validation allowed us to maintain comparable regime in terms of numbers of entities with the test set of other corpus. If cross-validation is known to present a risk of overfitting [25], we did not tuned the hyperparameters for the hybrid systems, hence the performance gain relative to the biGRU-CRF is a lower bound. Finally, in regards of the average low performance of the supervised model on APcNER (average F-measure of 67.1%), one could think the corpus unfit to allow for supervised learning. However, it still constitutes an important tool to evaluate semi-supervised or unsupervised systems. Combined with a more focused dataset (such as MERLOT), it could allow interesting transfer learning approaches to be tested.

# Conclusion

APcNER is a French corpus for clinical named-entity recognition of five type of entities which covers a large variety of document types. Extending supervised model with terminology allows for an easy performance gain, especially in low regimes of entities.

## Acknowledgements

We thank Nicolas Griffon for taking part in the annotation process.

## Authors' contributions

IL contributed to conceptualization; data curation; formal analysis; methodology; software; writing - original draft. NP contributed to conceptualization; data curation; formal analysis; software ; resources; writing - review & editing. XT contributed to conceptualization; data curation; formal analysis; methodology; project administration; resources; supervision; validation; writing - original draft; writing - review & editing.

## List of abbreviations

NER: named entity recognition
biGRU-CRF: Bidirectional - Gated Recurrent Unit - Conditional Random Field

AP-HP:Assistance Publique - Hôpitaux de Paris
IOB: inside outside beginning
ATC: Anatomical Therapeutic Chemical Classification System
BPDM: Base publique du médicament
CCAM: Classification commune des actes médicaux
CIM-10: Classification internationale des maladies
DRC: Dictionnaire des Resultats de Consultation
SNOMED: SYSTEMATIZED NOMENCLATURE OF MEDICINE CLINICAL TERMS

## DECLARATIONS

**Availability of data and material:** The code made for simulations will be made available.
**Competing interests**: The authors declare that they have no competing interests.